\documentclass[10pt,letterpaper]{article}

\usepackage{cogsci}
\usepackage{pslatex}
\usepackage{apacite}
\usepackage{graphicx}
\usepackage{subcaption}
\graphicspath{ {images/} }

\usepackage{fancyhdr}

\cogscifinalcopy 

\usepackage{pslatex}
\usepackage{float} 

\title{Linking Theories and Methods in Cognitive Sciences via Joint Embedding of the Scientific Literature: The Example of Cognitive Control}

\author{\bf{Morteza Ansarinia} \\ University of Luxembourg \\ Department of Behavioural and Cognitive Sciences \\ \texttt{morteza.ansarinia@uni.lu}
        \And
        \bf{Paul Schrater} \\ University of Minnesota, Twin Cities \\ Depts. of Computer Science and Psychology \\ \texttt{schrater@umn.edu}
        \AND
        \bf{Pedro Cardoso-Leite} \\ University of Luxembourg \\  Department of Behavioural and Cognitive Sciences \\ 
        \texttt{(pedro.cardosoleite@uni.lu)}}

\begin{document}

\maketitle

\begin{abstract}
Traditionally, theory and practice of Cognitive Control are linked via literature reviews by human domain experts. This approach, however, is inadequate to track the ever-growing literature. It may also be biased, and yield redundancies and confusion.

Here we present an alternative approach. We performed automated text analyses on a large body of scientific texts to create a joint representation of tasks and constructs. More specifically, 385,705 scientific abstracts were first mapped into an embedding space using a transformers-based language model. Document embeddings were then used to identify a task-construct graph embedding that grounds constructs \textit{on} tasks and supports nuanced meaning of the constructs by taking advantage of constrained random walks in the graph.

This joint task-construct graph embedding, can be queried to generate task batteries targeting specific constructs, may reveal knowledge gaps in the literature, and inspire new tasks and novel hypotheses.


\end{abstract}

\section{Introduction}
A key challenge in cognitive sciences, and in particular cognitive psychology and neuroscience, is to make sense of observable phenomena (i.e., behavior) in terms of theoretical constructs. Consider for instance Cognitive Control (CC)—a broad construct that comprises many components and engages multiple mechanisms which collectively aim to describe goal-directed behavior in a complex, uncertain world. 

CC is a major construct in cognitive sciences: In the year 2021 alone, PubMed indexed 974 papers with the term "Cognitive Control" in the title or abstract—an average of 3 papers per day. To understand CC, researchers have introduced a variety of theoretical constructs and conceived numerous cognitive tasks (see \citeNP{@baggetta2016}). However, the relationships between and within related constructs and tasks are not always clear. For example, because they are "measured" using the same set of tasks (e.g., Stroop, N-back, Digit Span, Stop-Signal, Task Switching), it seems reasonable to assume that cognitive control \cite{@botvinick2014}, executive functions \cite{@baggetta2016}, attentional control \cite{@reymermet2021}, and self-regulation \cite{@enkavi2019} are somewhat equivalent constructs; yet, they are not widely considered equal \cite{@nigg2016}.

Traditionally, the meaning and relationships between constructs and tasks are conceptualized in extensive literature reviews conducted by human experts. In this approach, researchers "manually" read, synthesize, and criticize the literature and write reviews or reports describing their understanding. Following such reviews, CC is viewed as interactions between generic core processes (e.g., inhibition, flexibility, working memory, and interference control in \citeNP{@diamond2013}), interactive componential \cite{@badre2011}, tasks-specific processes driven by goals \cite{@logan2017,@doebel2020}, or optimal parameterization of naturalistic tasks \cite{@botvinick2014}. This approach has been invaluable but it may also yield biased results \cite{@brick2021,@beam2021} and seems inadequate to track the ever-growing literature and stay current. In this context, modern machine learning methods may provide useful and complementary insights.

When considering terms in the literature, there are two major impediments to creating consistent construct-task associations: \textit{construct hypernomy} when conceptualizing CC and \textit{task impurity} when operationalizing it. \textit{Construct hypernomy} occurs when description of the same construct varies across different contexts due to the way it is assessed. It creates different meanings of the same concept. "Attentional Control", for example, likely means something different in \citeA{@ahissar1993} (as measured by low-level perceptual tasks) than it does in \citeA{@burgoyne2020} (as measured by complex cognitive tasks). \textit{Task impurity}, on the other hand, refers to the idea that performance on a task loads onto multiple constructs (i.e., there is not a one-to-one mapping between constructs and tasks). Because of the impurity, no task taps into just one isolated construct. Performance in the Backward Digit Span, for instance, involves short-term memory, visual perception, sustained attention and working memory, to name just a few. The consequence is that constructs lack a consistent, groundable semantic content, corrupting interpretations of neural and cognitive research that depend on them.

Construct hypernomy and task impurity are quite common in CC research because complex concepts like cognitive control manifest themselves differently across different individuals and contexts \cite{@burgoyne2020}. For that, researchers often use multiple tasks in their studies and apply statistical methods such as latent factor analysis to discern underlying constructs. Nevertheless, the resulting latent models of CC are rarely agreed upon, as is the selection of tasks (see, for example, \citeNP{@reymermet2021}; \citeNP{@doebel2020}, \citeNP{@enkavi2019}, \citeNP{@nigg2016}).

Ambiguous associations of constructs and tasks make it hard to interpret past results, hinder scientific progress and the development of effective interventions. With the advent of scalable machine learning, however, construct-task associations may be clarified. The goal of this paper is to approach the conceptual richness of a large body of scientific works and take advantage of recent context-aware language models in machine learning to clarify the association of CC tasks and constructs. More specifically, we collect and analyze scientific texts about CC tasks and constructs and encode text data into rich semantic embeddings using transfer learning. Transfer learning exploits the rich representations generated by natural language models trained to faithfully represent contextual meaning—unlike traditional bag-of-word or clustering techniques. Similarities between embedded representations are then used to build up a hypergraph \cite{@battiston2021} that connects tasks and constructs.

First, we show that this hypergraph representation regrounds constructs on tasks and provides nuanced meaning of the constructs, ultimately demonstrating construct hypernomy. Second we show that pulling theoretical and experimental literature into overlapping components of a hypergraph may greatly benefit researchers: the joint task-construct embeddings can be queried to generate special-purpose task batteries, it may reveal knowledge gaps, inspire the design of new experiments and yield novel hypotheses regarding the structure and function of CC. This empirical and descriptive model of the literature, rather than expert-driven ones, may also be used in future applications to enhance knowledge searches (see \citeNP{@beam2021} for a comparison of a data-driven mapping of the literature and expert-driven knowledge frameworks like DSM and RDoC).

\section{Methods}

{\bf Data.} We created a lexicon of CC-related terms (172 terms, of which 72 were task names and 100 were construct names) based on the previously published work on Cognitive Control \cite{@barch2009}, Attentional Control \cite{@bastian2021}, Executive Functions \cite{@baggetta2016, @diamond2013}, and Self Regulation \cite{@enkavi2019}. Each term in the lexion was associated with a PubMed-specific search query by which papers with the term in their title or abstract were retrieved. This resulted in a dataset of loosely labeled documents, each labeled by one or more lexicon terms (n=522,972 hits, of which 385,705 were unique). For the purpose of the current analyses we only retained the title and abstract of the papers, along with the lexicon terms that were used to retrieve them. Having multiple labels per document was crucial to quantify the co-appearance of the terms in the literature. After the documents were collected we removed 14 terms from the lexicon because they yielded too few documents to support cross validation splits ($n<5$).

{\bf Analysis.} To understand the relationships among and between tasks and constructs, our goal is to build graphs that represent tasks and constructs as nodes and measure similarity/distance between them as edges. Graph $G$ can be used to jointly infer embeddings of both construct and task nodes in a shared vector space, in that relative closeness of two nodes is estimated by the similarities of node attributes as well as the shared neighbors in the graph. Heterogeneous graph $G=(V_{tasks} \bigcup V_{constructs}, E)$ is defined by its two types of nodes, $V_{tasks}$ and $V_{constructs}$, labeled by either a task or a construct term, while the weighted edges, $E$, represent the links between two or more nodes, reflecting similarity of the corresponding terms in the literature. Node attributes being relevant scientific texts, the existence and weight of a link between two nodes is predicted by the similarity of corresponding node attributes; the higher the similarity between node attributes, the higher the chance of the nodes being associated. The core problem becomes learning task and construct attribute embeddings that predict co-occurrence and semantic similarity measures. We used the following steps to create the graph $G$ from the collected scientific texts.

\begin{figure*}[htp]
\subcaptionbox*{(A)}{\includegraphics[width=4.3cm]{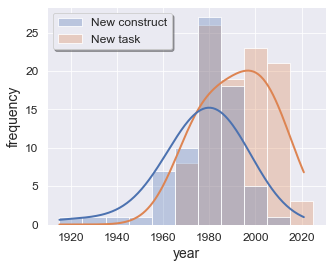}}
\hfill
\subcaptionbox*{(B)}{\includegraphics[width=4.3cm]{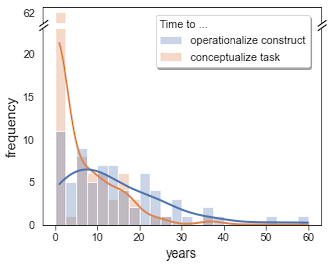}}
\subcaptionbox*{(C)}{\includegraphics[width=4.3cm]{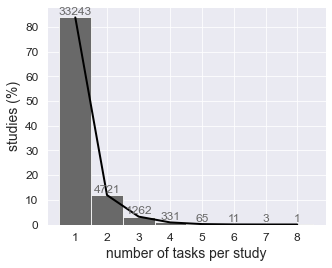}}
\subcaptionbox*{(D)}{\includegraphics[width=4.3cm]{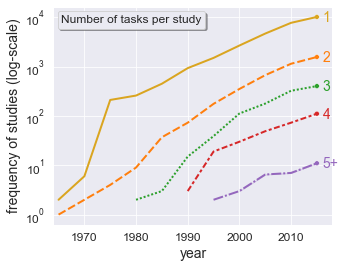}}
\caption{(A) Introducing new tasks (task innovation) and constructs (concept innovation) is characterized by a burst followed by declining innovation. (B) Task and Construct occurrences in publication abstracts are temporally decoupled. Time to operationalize constructs (blue) is the time between the first occurrence of a construct and the first co-occurrence of that construct with any tasks, while Time to conceptualize tasks (orange) is the time between the first occurrence of a task and the first co-occurrence of that task with any of the construct. (C) The majority of the literature only used one task in their studies, showing a lack of multitask design of experiments. (D) While the number of papers published each year increases exponentially, the number of tasks per study remains fairly constant across time.} 
\label{figure-1}
\end{figure*}

\begin{figure*}[ht]
\begin{center}
\includegraphics[width=18cm]{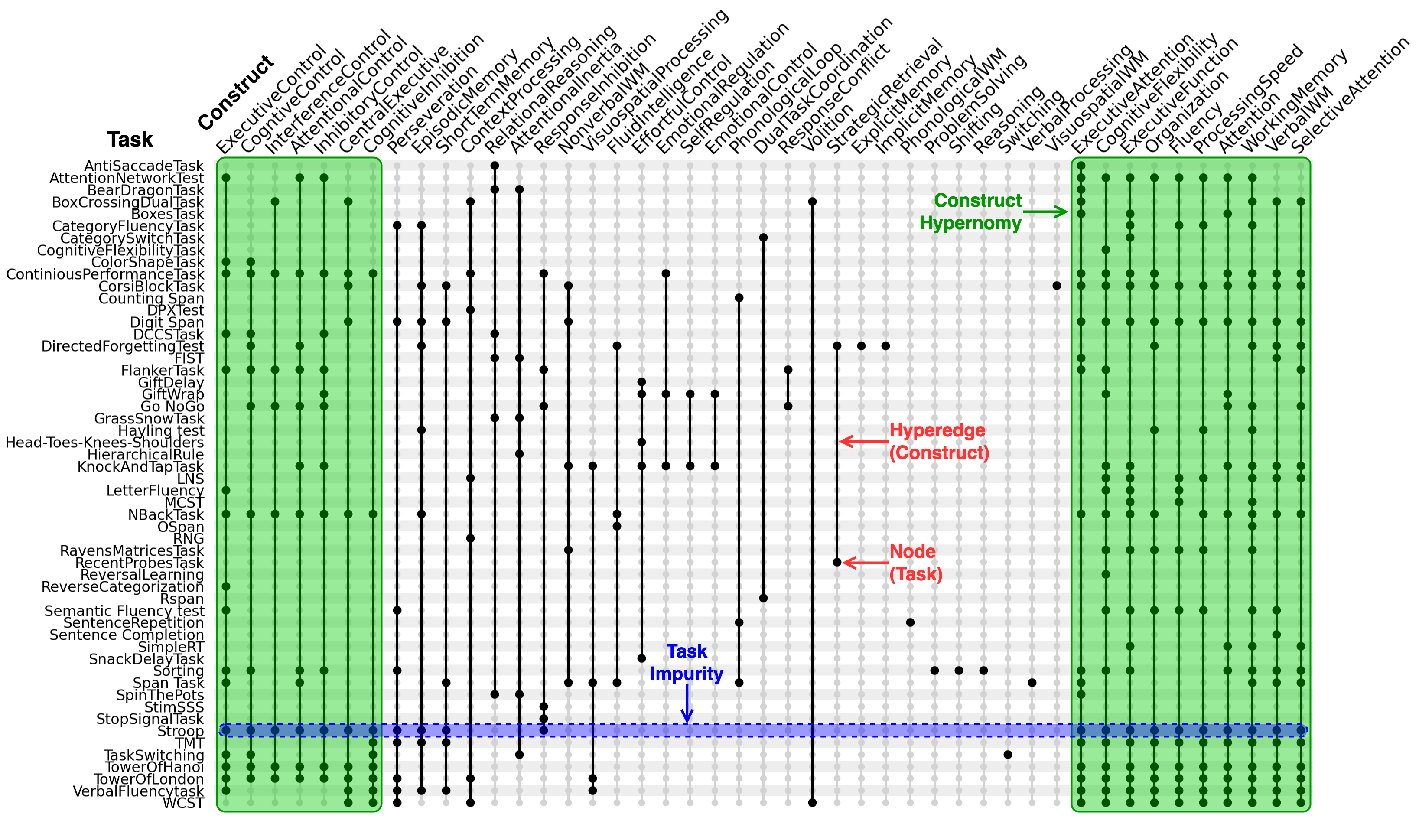}
\end{center}
\caption{Task-Construct hypergraph: representations of control-related constructs as hyperedges (vertical black lines) over a subset of tasks (nodes). Construct hypernomy is reflected as overlapping hyperedges (e.g., green regions), and task impurity as nodes scattered over multiple hyperedges (e.g., blue region). Distances between nodes are not meaningful. Nodes are reorganized for visual clarity and only a subset of the graph is displayed.}
\label{figure-2}
\end{figure*}

The data collection resulted in a dataset of 385,705 unique, but loosely-labeled, abstract texts, all of which were then encoded into embeddings of 1024 dimensions using a pre-trained transformer language model (GPT-3 Ada for text similarity embedding; see \citeNP{@brown2020}). The language model transformed raw texts into 1024-dimensional vectors, gpt3-embedding, representing semantic similarity between two or more pieces of text. Since keeping the original structure of the text was important for the model to understand the context, we did not preprocess the raw text. To convert text similarity into a shared topic representation (which improves relating task and construct text embeddings), we applied Top2Vec topic modeling \cite{@angelov2020} to the gpt3-embedding which projected them into a space of 473 dimensions, i.e., topic-embeddings. Each column of the topic-embedding matrix represents a topic, and element $ij$ shows the probability of assigning document $i$ to the topic $j$. Realigning the gpt3-embedding into topic-embeddings improved the quality of the dataset for a number of reasons. First, it improves the quality of the labels in the dataset by discarding outlier documents. These are documents that belong to no topics of interest or are assigned to irrelevant topics (e.g., genetics)—after removing outlier documents, 293'014 unique documents remained for further analysis. Second, topic modeling allows one to extract a useful, interpretable representation of the documents, as each dimension of the topic-embedding shows the probability of assigning a document to a topic while being faithful to the contextual representation of the documents in the gpt3-embedding space. This generates a digraph between nodes representing lexicon terms and the topic-embedding vectors.

To convert this into a construct-task graph, we grouped lexical terms associated with construct and tasks to generate graph nodes. To compute topic-similarity between groups of lexical terms associated with each construct or task node, we fitted a multivariate normal distribution over the topic vectors  of each node separately and then calculated the distance between all nodes as measured by the Jensen-Shannon divergence of those node-level distributions. This step added edges to the graph, G, with edges weighted by the inverse distance of nodes in the JS-divergence matrix.

To learn a representation of the graph that only preserves paths from tasks to constructs and vice versa, we then applied Metapath2Vec (1000 random walks of step size 100, accompanied by skip-gram Word2Vec embedding of size 128 and maximum window size of 5; as recommended in \citeNP{@ruch2020}). The Metapath2Vec embedding encodes random walks of specific patterns in a heterogeneous graph, here patterns being alternating random walks between task and construct nodes.

Finally, by applying HDBSCAN soft clustering to the node attributes and thresholding the edges (discarding all the edges weighed within one standard deviation from median), we transform the graph G to a homogeneous hypergraph, i.e., nodes are now only of type task, while constructs are hyperedges that group a subset of tasks in overlapping clusters.

\section{Results}

We used a variety of data-driven approaches to collect and understand CC publications. Briefly, we a) created an all-inclusive lexion of construct and task terms, b) queried PubMed to collect relevant abstract texts, c) vectorized all the raw texts using GPT-3 Sentence Similarity Embedding, an unsupervised pre-trained language model, d) applied Top2Vec topic modeling technique to all the document embeddings together and identified dimensions of a useful latent space, i.e., topics. We then created a graphical representation of the lexicon terms, i.e., task-construct graph, and used them to predict the association between terms.

{\bf The richness of tasks and constructs in the literature.} Although there are many task and construct terms, their relative frequencies differ widely. For example, "Stroop Task" is mentioned 8,003 times in the period 1973--2022 while "Delay Discounting Task" was only mentioned 466 times over the same period of time. The use of each term tends to increase over time. Interestingly the rate at which new constructs and tasks are introduced does not follow the same curve as the number of publications in the field; rather there seems to have been a peak of innovation for constructs around 1980 and for tasks around the year 2000 (Figure 1A). Such patterns, visible in simple descriptive statistics (Figure 1), may provide interesting insights into understanding the maturity and vitality of a research field.

{\bf Regrounding constructs on tasks.} It took on average 7 years for the constructs to be explicitly associated with a task (see Figure 1B). The meaning of a theoretical construct may change across time and gain clarity and precision with new empirical measures and cognitive tasks being used by the research community to flesh out the construct. A core idea in this paper is that by evaluating how constructs are operationalized (i.e., linked to cognitive tasks) key insight can be gained about what a construct means. Grounding the definition of constructs on tasks provides a nuanced meaning of constructs that relies on observable measures. It also allows the computation of useful measures on constructs (e.g., specificity) and on between pairs of constructs (e.g., measures of redundancy, similarity, and distance). To investigate the relationships between cognitive constructs, we use hyperedges in the task-construct graph as a measure of similarity, indicating the extent to which a construct hyperedge can be reconstructed by neighboring tasks.

{\bf Construct hypernomy.} The task-construct graph readily demonstrates construct hypernomy and task impurity in the CC literature. We first sought hypernomy as highly overlapping hyperedges of seemingly incompatible constructs, as well as a high degree of task nodes with neighboring constructs as a measure of the task impurity. Figure 2 illustrates overlapping hyperedges of the most popular constructs where hyperedges for Cognitive Control, Executive Control, Behavioral Control, Central Executive, and Attentional Control are overlapping and identical.

\begin{figure*}[ht]
\begin{center}
\includegraphics[width=18cm]{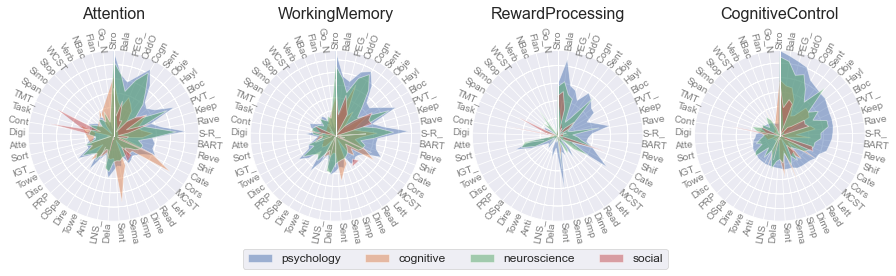}
\end{center}
\caption{Associations between tasks and constructs minimally overlap across scientific disciplines. Rose plots show the relative association between constructs and tasks, with each color representing a different field. Lack of overlap between the "spikes" indicates disjoint operationalizations across fields.} 
\label{figure-3}
\end{figure*}

\begin{figure}[ht]
\includegraphics[width=8.5cm]{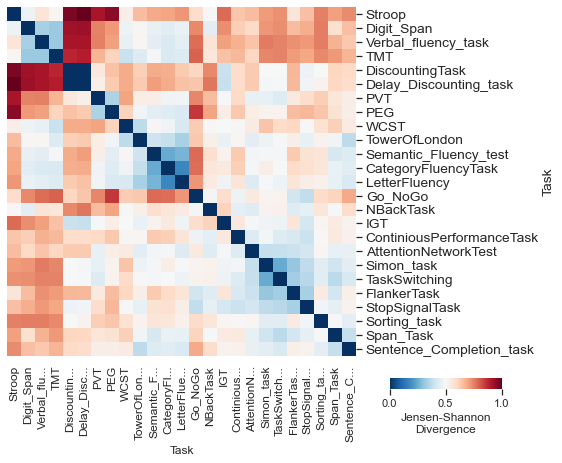}
\caption{Pairwise distances between the 25 most popular cognitive control tasks as measured by the symmetric Jensen-Shannon divergence of two multivariate normal distributions of their node attributes in the task-construct graph. Higher divergence indicates higher dissimilarity between corresponding scientific texts. Task-task distances may for example provide a data-driven proxy for predicting and explaining transfer effects in cognitive training research.} 
\label{figure-4}
\end{figure}

{\bf Task inconsistency across disciplines.} A major source of hypernomy stems from descriptions and measurements of the constructs often being inconsistent across scientific communities. To test this idea, we sought to determine whether construct hyperedges, and their task associations, vary across four cognitive disciplines (psychology, neuroscience, cognitive science, and social science). Using the same method described in the analysis section we created four discipline-specific graph embeddings. The only difference was that publications were grouped by discipline, which was determined by searching for the terms "social", "psycho", "neur", or "cognit" in the journal titles. Constructs that have inconsistent task associations across the disciplines are hypernomic (Figure 3).

{\bf Refactoring tasks and constructs.} Designing effective assessments of CC can be challenging for a number of reasons. Participants have limited time to spend on cognitive tasks. 1) If these tasks are poorly selected, performance on these tasks may not be very informative (e.g., measures are conceptually redundant); 2) If only one task is used, the inferential resolution of performance to construct is very limited. Thus in order to be able to make specific theoretical claims about CC it is necessary to use multiple, well-chosen tasks in experiments. This is currently not the case. As shown in Figure 1C, most research uses only one task. In fact, only 17 percent of publications used 2 or more tasks. The task-construct graph presented here may facilitate novel experimental designs of such multi-task, max-information experiments by providing a similarity-based space in which tasks can be identified, and grouped, by the overlapping subgraphs (i.e., constructs) that they belong to.

In the task-construct graph, two tasks are similar if they share identical neighbors, i.e., constructs. And tasks cover a set of constructs if their union set overlaps the corresponding hyperedges of the constructs. These principles equip researchers with sound and quantified methods to refactor tasks (e.g., discard redundant tasks, quantitatively measuring similarity of tasks via constructs, and performing set operations on a group of tasks). Such a refactored set of tasks controls the construct-redundancy of tasks and will shorten the time required to complete comprehensive assessments. It provides a method to design a task battery to effectively cover constructs (i.e., minimal redundancy while measuring different facets of the constructs).

{\bf Sparsity in the task space.} There are numerous cognitive tasks in the literature; how these tasks relate to each other remains unclear. There are many cognitive control tasks that are rarely used (see \citeNP{@baggetta2016}), and even fewer used in combination with other tasks. Even when tasks were used together, their relationship might still be unclear. The question of how tasks relate to each other is key in the cognitive training domain where researchers aim to train cognitive abilities in general rather than performance on a specific task. In that context, a common point of disagreement is to predict and interpret transfer effects (i.e., how much training in task A improves performance in task B). A measure of distance between tasks based on their grounding on constructs may provide an objective foundation to understand these transfer effects—the task-construct graph embedding proposed here provides a means to compute such inter-task distances.

To quantify the distance between two cognitive tasks, we compute the Jensen-Shannon divergence between their node embeddings in the task-construct graph. Figure 4 shows, for example, that the Trail Making Task is relatively close to the Digit Span Task, suggesting its training effects transfer more easily to the Digit Span Task than to tasks such as the Discounting Task.

Distance between the task nodes can also allow us to identify gaps in the task space: gaps may be visible as disconnected graph components. Identifying such gaps may reveal opportunities to develop new useful tasks. Alternatively, there may only exist associations between groups of tasks and groups of constructs—i.e. the task-construct associations are not \textit{atomic}. This reflects a lack of purity in the tasks or constructs or both that might be improved by refactoring constructs or decomposing tasks into components. 

{\bf Querying the graph embedding for task batteries targeting specific cognitive constructs.} Some studies use batteries of tasks that together address a research question and measure one or more constructs from several viewpoints. The process of building such task batteries can be facilitated by leveraging the task-construct graph embedding; one can query the graph for an array of tasks spanning a given set of constructs. The joint embedding translates queries into arithmetic operations in the embedding space (positive samples and negative samples), allowing for more explicit and visible decisions.

Query operations on the task-construct graph are made possible by using the underlying node embedding vectors extracted as a part of Metapath2Vec graph embedding. Queries include, for example, prioritizing tasks for a given construct, or a set of tasks for a set of constructs. To prioritize tasks for a construct, the task-construct graph looks for task nodes that are closest to the simple mean of the queried construct, e.g., in terms of sum of weighted node embeddings. And for a list of tasks for multiple constructs, find the minimum spanning tree that covers all the queried construct hyperedges. For example, if one queries \texttt{(Reward Processing + ReversalLearning - GoNoGo - SortingTask)}, one will get the recommendation to use the \texttt{BART}, \texttt{GiftDelay}, \texttt{BalanceBeam} \cite{@baggetta2016}, and \texttt{StimSSS} \cite{@enkavi2019} tasks, which are ordered by the cosine similarity between the mean vector of the query and the task vectors in the graph embedding model.

\section{Implications}

Ambiguous meanings and relationships between cognitive tasks and constructs call for a more rigorous way to handle constructs—an obvious solution would be to adopt a more formal notation and refer to specific knowledge models (e.g., ontologies). The knowledge model must be flexible enough to capture a wide range of association between constructs and tasks. The proposed task-construct graph embedding provides a useful representation of the cognitive control literature built upon topic embedding. In this representation, association of two entities, e.g., task-construct, relies on shared topics as well as the walks between them in a graph representation. By predicting links using topic embeddings of the nodes, we find most similar aspects of, for example, two constructs, a similarity that could be explainable in natural language.

A consistent, sound, and parsimonious framework of CC has been desired from the beginning. Yet, the growing number of publications and newly introduced constructs makes it impossible to integrate them into a bigger picture. While researchers may disagree on theoretical perspectives and thus on which terms to use, they generally might agree on the fact that if two constructs are "measured" by the same tasks, the constructs must be somewhat related. We proposed a joint embedding of constructs and tasks (based on scientific texts in a graph representation) to drive a more nuanced interpretation of the constructs by regrounding abstract constructs on the concrete set of observable tasks.

The proposed graph-based embedding enables explanatory reasoning driven by scientific texts. Unlike expert-driven models, the models reason regardless of the preferences in research; yet it is not clear whether other kinds of biases are addressed as the knowledge source and pre-trained language model are themselves produced by humans. By scaling up the knowledge model to a large body of available texts, the model is able to encapsulate even more aspects of Cognitive Control, and in general, multidisciplinary research.

Disagreements about the meaning of a construct are partly explained by differences in how we interpret responses to a particular task. By focusing on the co-occurrence of task and construct names in scientific texts, our approach implicitly makes strong assumptions about the relationship between abstract constructs and their imperfect but observable measures. The limitations of the present work can be partially addressed by expanding the hypergraph to include, for example, concepts such as brain mechanisms, research communities, and analysis techniques.

Explainable symbolic AI and machine learning have been long in debate to model knowledge. Regardless of the specific topic discussed here (i.e., Cognitive Control), the proposed model can be seen as an effort to connect symbolic modeling (as in ontologies) and machine learning (as in embeddings). Our method informs an ontology of scientific texts using context-aware embeddings that are extracted from a loosely-labeled body of scientific texts requiring minimal human input. It is an automated pipeline that only requires a lexicon, builds on large-scale language models and that can scale to millions of documents, making it a viable approach to meaningfully monitor the scientific literature continuously and extensively.

\section{Acknowledgments}

This research was supported by the Luxembourg National Research Fund (ATTRACT/2016/ID/11242114/DIGILEARN) and (INTER Mobility/2017-2/ID/11765868/ULALA).

\bibliographystyle{apacite}

\setlength{\bibleftmargin}{.125in}
\setlength{\bibindent}{-\bibleftmargin}

\bibliography{cogsci2022_cogtext}

\begin{thebibliography}{}

\bibitem [\protect \citeauthoryear {%
Ahissar%
\ \BBA {} Hochstein%
}{%
Ahissar%
\ \BBA {} Hochstein%
}{%
{\protect \APACyear {1993}}%
}]{%
@ahissar1993}
\APACinsertmetastar {%
@ahissar1993}%
\begin{APACrefauthors}%
Ahissar, M.%
\BCBT {}\ \BBA {} Hochstein, S.%
\end{APACrefauthors}%
\unskip\
\newblock
\APACrefYearMonthDay{1993}{}{}.
\newblock
{\BBOQ}\APACrefatitle {{Attentional control of early perceptual learning}}
  {{Attentional control of early perceptual learning}}.{\BBCQ}
\newblock
\APACjournalVolNumPages{Proceedings of the National Academy of
  Sciences}{90}{12}{}.
\PrintBackRefs{\CurrentBib}

\bibitem [\protect \citeauthoryear {%
Angelov%
}{%
Angelov%
}{%
{\protect \APACyear {2020}}%
}]{%
@angelov2020}
\APACinsertmetastar {%
@angelov2020}%
\begin{APACrefauthors}%
Angelov, D.%
\end{APACrefauthors}%
\unskip\
\newblock
\APACrefYearMonthDay{2020}{}{}.
\newblock
{\BBOQ}\APACrefatitle {{Top2Vec: Distributed Representations of Topics}}
  {{Top2Vec: Distributed Representations of Topics}}.{\BBCQ}
\newblock
\APACjournalVolNumPages{arXiv}{}{}{}.
\PrintBackRefs{\CurrentBib}

\bibitem [\protect \citeauthoryear {%
Badre%
}{%
Badre%
}{%
{\protect \APACyear {2011}}%
}]{%
@badre2011}
\APACinsertmetastar {%
@badre2011}%
\begin{APACrefauthors}%
Badre, D.%
\end{APACrefauthors}%
\unskip\
\newblock
\APACrefYearMonthDay{2011}{}{}.
\newblock
{\BBOQ}\APACrefatitle {{Defining an Ontology of Cognitive Control Requires
  Attention to Component Interactions}} {{Defining an Ontology of Cognitive
  Control Requires Attention to Component Interactions}}.{\BBCQ}
\newblock
\APACjournalVolNumPages{Topics in Cognitive Science}{3}{2}{}.
\PrintBackRefs{\CurrentBib}

\bibitem [\protect \citeauthoryear {%
Baggetta%
\ \BBA {} Alexander%
}{%
Baggetta%
\ \BBA {} Alexander%
}{%
{\protect \APACyear {2016}}%
}]{%
@baggetta2016}
\APACinsertmetastar {%
@baggetta2016}%
\begin{APACrefauthors}%
Baggetta, P.%
\BCBT {}\ \BBA {} Alexander, P\BPBI A.%
\end{APACrefauthors}%
\unskip\
\newblock
\APACrefYearMonthDay{2016}{}{}.
\newblock
{\BBOQ}\APACrefatitle {{Conceptualization and Operationalization of Executive
  Function}} {{Conceptualization and Operationalization of Executive
  Function}}.{\BBCQ}
\newblock
\APACjournalVolNumPages{Mind, Brain, and Education}{10}{1}{}.
\PrintBackRefs{\CurrentBib}

\bibitem [\protect \citeauthoryear {%
Barch%
, Braver%
, Carter%
, Poldrack%
\BCBL {}\ \BBA {} Robbins%
}{%
Barch%
\ \protect \BOthers {.}}{%
{\protect \APACyear {2009}}%
}]{%
@barch2009}
\APACinsertmetastar {%
@barch2009}%
\begin{APACrefauthors}%
Barch, D\BPBI M.%
, Braver, T\BPBI S.%
, Carter, C\BPBI S.%
, Poldrack, R\BPBI A.%
\BCBL {}\ \BBA {} Robbins, T\BPBI W.%
\end{APACrefauthors}%
\unskip\
\newblock
\APACrefYearMonthDay{2009}{}{}.
\newblock
{\BBOQ}\APACrefatitle {{CNTRICS Final Task Selection: Executive Control}}
  {{CNTRICS Final Task Selection: Executive Control}}.{\BBCQ}
\newblock
\APACjournalVolNumPages{Schizophrenia Bulletin}{35}{1}{}.
\PrintBackRefs{\CurrentBib}

\bibitem [\protect \citeauthoryear {%
Bastian%
\ \protect \BOthers {.}}{%
Bastian%
\ \protect \BOthers {.}}{%
{\protect \APACyear {{\protect \bibnodate {}}}}%
}]{%
@bastian2021}
\APACinsertmetastar {%
@bastian2021}%
\begin{APACrefauthors}%
Bastian, C\BPBI C\BPBI v.%
, Blais, C.%
, Brewer, G\BPBI A.%
, Gyurkovics, M.%
, Hedge, C.%
, Kałamała, P.%
\BDBL {}Wiemers, E\BPBI A.%
\end{APACrefauthors}%
\unskip\
\newblock
\APACrefYearMonthDay{{\protect \bibnodate {}}}{}{}.
\newblock
{\BBOQ}\APACrefatitle {{Advancing the understanding of individual differences
  in attentional control: Theoretical, methodological, and analytical
  considerations}} {{Advancing the understanding of individual differences in
  attentional control: Theoretical, methodological, and analytical
  considerations}}.{\BBCQ}
\newblock
\APACjournalVolNumPages{PsyArXiv}{}{}{}.
\newblock
\begin{APACrefDOI} \doi{10.31234/osf.io/x3b9k} \end{APACrefDOI}
\PrintBackRefs{\CurrentBib}

\bibitem [\protect \citeauthoryear {%
Battiston%
\ \protect \BOthers {.}}{%
Battiston%
\ \protect \BOthers {.}}{%
{\protect \APACyear {2021}}%
}]{%
@battiston2021}
\APACinsertmetastar {%
@battiston2021}%
\begin{APACrefauthors}%
Battiston, F.%
, Amico, E.%
, Barrat, A.%
, Bianconi, G.%
, Arruda, G\BPBI F\BPBI d.%
, Franceschiello, B.%
\BDBL {}Petri, G.%
\end{APACrefauthors}%
\unskip\
\newblock
\APACrefYearMonthDay{2021}{}{}.
\newblock
{\BBOQ}\APACrefatitle {{The physics of higher-order interactions in complex
  systems}} {{The physics of higher-order interactions in complex
  systems}}.{\BBCQ}
\newblock
\APACjournalVolNumPages{Nature Physics}{17}{10}{}.
\PrintBackRefs{\CurrentBib}

\bibitem [\protect \citeauthoryear {%
Beam%
, Potts%
, Poldrack%
\BCBL {}\ \BBA {} Etkin%
}{%
Beam%
\ \protect \BOthers {.}}{%
{\protect \APACyear {2021}}%
}]{%
@beam2021}
\APACinsertmetastar {%
@beam2021}%
\begin{APACrefauthors}%
Beam, E.%
, Potts, C.%
, Poldrack, R\BPBI A.%
\BCBL {}\ \BBA {} Etkin, A.%
\end{APACrefauthors}%
\unskip\
\newblock
\APACrefYearMonthDay{2021}{}{}.
\newblock
{\BBOQ}\APACrefatitle {{A data-driven framework for mapping domains of human
  neurobiology}} {{A data-driven framework for mapping domains of human
  neurobiology}}.{\BBCQ}
\newblock
\APACjournalVolNumPages{Nature Neuroscience}{}{}{}.
\PrintBackRefs{\CurrentBib}

\bibitem [\protect \citeauthoryear {%
Botvinick%
\ \BBA {} Cohen%
}{%
Botvinick%
\ \BBA {} Cohen%
}{%
{\protect \APACyear {2014}}%
}]{%
@botvinick2014}
\APACinsertmetastar {%
@botvinick2014}%
\begin{APACrefauthors}%
Botvinick, M\BPBI M.%
\BCBT {}\ \BBA {} Cohen, J\BPBI D.%
\end{APACrefauthors}%
\unskip\
\newblock
\APACrefYearMonthDay{2014}{}{}.
\newblock
{\BBOQ}\APACrefatitle {{The Computational and Neural Basis of Cognitive
  Control: Charted Territory and New Frontiers}} {{The Computational and Neural
  Basis of Cognitive Control: Charted Territory and New Frontiers}}.{\BBCQ}
\newblock
\APACjournalVolNumPages{Cognitive Science}{38}{6}{}.
\PrintBackRefs{\CurrentBib}

\bibitem [\protect \citeauthoryear {%
Brick%
, Hood%
, Ekroll%
\BCBL {}\ \BBA {} de Wit%
}{%
Brick%
\ \protect \BOthers {.}}{%
{\protect \APACyear {2021}}%
}]{%
@brick2021}
\APACinsertmetastar {%
@brick2021}%
\begin{APACrefauthors}%
Brick, C.%
, Hood, B.%
, Ekroll, V.%
\BCBL {}\ \BBA {} de Wit, L.%
\end{APACrefauthors}%
\unskip\
\newblock
\APACrefYearMonthDay{2021}{}{}.
\newblock
{\BBOQ}\APACrefatitle {{Illusory Essences: A Bias Holding Back Theorizing in
  Psychological Science}} {{Illusory Essences: A Bias Holding Back Theorizing
  in Psychological Science}}.{\BBCQ}
\newblock
\APACjournalVolNumPages{Perspectives on Psychological Science}{}{}{}.
\PrintBackRefs{\CurrentBib}

\bibitem [\protect \citeauthoryear {%
Brown%
\ \protect \BOthers {.}}{%
Brown%
\ \protect \BOthers {.}}{%
{\protect \APACyear {2020}}%
}]{%
@brown2020}
\APACinsertmetastar {%
@brown2020}%
\begin{APACrefauthors}%
Brown, T\BPBI B.%
, Mann, B.%
, Ryder, N.%
, Subbiah, M.%
, Kaplan, J.%
, Dhariwal, P.%
\BDBL {}Amodei, D.%
\end{APACrefauthors}%
\unskip\
\newblock
\APACrefYearMonthDay{2020}{}{}.
\newblock
{\BBOQ}\APACrefatitle {{Language Models are Few-Shot Learners}} {{Language
  Models are Few-Shot Learners}}.{\BBCQ}
\newblock
\APACjournalVolNumPages{arXiv}{}{}{}.
\PrintBackRefs{\CurrentBib}

\bibitem [\protect \citeauthoryear {%
Burgoyne%
\ \BBA {} Engle%
}{%
Burgoyne%
\ \BBA {} Engle%
}{%
{\protect \APACyear {2020}}%
}]{%
@burgoyne2020}
\APACinsertmetastar {%
@burgoyne2020}%
\begin{APACrefauthors}%
Burgoyne, A\BPBI P.%
\BCBT {}\ \BBA {} Engle, R\BPBI W.%
\end{APACrefauthors}%
\unskip\
\newblock
\APACrefYearMonthDay{2020}{}{}.
\newblock
{\BBOQ}\APACrefatitle {{Attention Control: A Cornerstone of Higher-Order
  Cognition}} {{Attention Control: A Cornerstone of Higher-Order
  Cognition}}.{\BBCQ}
\newblock
\APACjournalVolNumPages{Current Directions in Psychological Science}{29}{6}{}.
\PrintBackRefs{\CurrentBib}

\bibitem [\protect \citeauthoryear {%
Diamond%
}{%
Diamond%
}{%
{\protect \APACyear {2013}}%
}]{%
@diamond2013}
\APACinsertmetastar {%
@diamond2013}%
\begin{APACrefauthors}%
Diamond, A.%
\end{APACrefauthors}%
\unskip\
\newblock
\APACrefYearMonthDay{2013}{}{}.
\newblock
{\BBOQ}\APACrefatitle {{Executive Functions}} {{Executive Functions}}.{\BBCQ}
\newblock
\APACjournalVolNumPages{Annual Review of Psychology}{64}{1}{}.
\PrintBackRefs{\CurrentBib}

\bibitem [\protect \citeauthoryear {%
Doebel%
}{%
Doebel%
}{%
{\protect \APACyear {2020}}%
}]{%
@doebel2020}
\APACinsertmetastar {%
@doebel2020}%
\begin{APACrefauthors}%
Doebel, S.%
\end{APACrefauthors}%
\unskip\
\newblock
\APACrefYearMonthDay{2020}{}{}.
\newblock
{\BBOQ}\APACrefatitle {{Rethinking Executive Function and Its Development}}
  {{Rethinking Executive Function and Its Development}}.{\BBCQ}
\newblock
\APACjournalVolNumPages{Perspectives on Psychological Science}{15}{4}{}.
\PrintBackRefs{\CurrentBib}

\bibitem [\protect \citeauthoryear {%
Enkavi%
\ \protect \BOthers {.}}{%
Enkavi%
\ \protect \BOthers {.}}{%
{\protect \APACyear {2019}}%
}]{%
@enkavi2019}
\APACinsertmetastar {%
@enkavi2019}%
\begin{APACrefauthors}%
Enkavi, A\BPBI Z.%
, Eisenberg, I\BPBI W.%
, Bissett, P\BPBI G.%
, Mazza, G\BPBI L.%
, MacKinnon, D\BPBI P.%
, Marsch, L\BPBI A.%
\BCBL {}\ \BBA {} Poldrack, R\BPBI A.%
\end{APACrefauthors}%
\unskip\
\newblock
\APACrefYearMonthDay{2019}{}{}.
\newblock
{\BBOQ}\APACrefatitle {{Large-scale analysis of test–retest reliabilities of
  self-regulation measures}} {{Large-scale analysis of test–retest
  reliabilities of self-regulation measures}}.{\BBCQ}
\newblock
\APACjournalVolNumPages{Proceedings of the National Academy of
  Sciences}{116}{12}{}.
\PrintBackRefs{\CurrentBib}

\bibitem [\protect \citeauthoryear {%
Logan%
}{%
Logan%
}{%
{\protect \APACyear {2017}}%
}]{%
@logan2017}
\APACinsertmetastar {%
@logan2017}%
\begin{APACrefauthors}%
Logan, G\BPBI D.%
\end{APACrefauthors}%
\unskip\
\newblock
\APACrefYearMonthDay{2017}{}{}.
\newblock
{\BBOQ}\APACrefatitle {{Taking Control of Cognition: An Instance Perspective on
  Acts of Control}} {{Taking Control of Cognition: An Instance Perspective on
  Acts of Control}}.{\BBCQ}
\newblock
\APACjournalVolNumPages{American Psychologist}{72}{9}{}.
\PrintBackRefs{\CurrentBib}

\bibitem [\protect \citeauthoryear {%
Nigg%
}{%
Nigg%
}{%
{\protect \APACyear {2016}}%
}]{%
@nigg2016}
\APACinsertmetastar {%
@nigg2016}%
\begin{APACrefauthors}%
Nigg, J\BPBI T.%
\end{APACrefauthors}%
\unskip\
\newblock
\APACrefYearMonthDay{2016}{}{}.
\newblock
{\BBOQ}\APACrefatitle {{Annual Research Review: On the relations among
  self-regulation, self-control, executive functioning, effortful control,
  cognitive control, impulsivity, risk-taking, and inhibition for developmental
  psychopathology}} {{Annual Research Review: On the relations among
  self-regulation, self-control, executive functioning, effortful control,
  cognitive control, impulsivity, risk-taking, and inhibition for developmental
  psychopathology}}.{\BBCQ}
\newblock
\APACjournalVolNumPages{Journal of Child Psychology and Psychiatry}{58}{4}{}.
\PrintBackRefs{\CurrentBib}

\bibitem [\protect \citeauthoryear {%
Rey-Mermet%
, Singmann%
\BCBL {}\ \BBA {} Oberauer%
}{%
Rey-Mermet%
\ \protect \BOthers {.}}{%
{\protect \APACyear {2021}}%
}]{%
@reymermet2021}
\APACinsertmetastar {%
@reymermet2021}%
\begin{APACrefauthors}%
Rey-Mermet, A.%
, Singmann, H.%
\BCBL {}\ \BBA {} Oberauer, K.%
\end{APACrefauthors}%
\unskip\
\newblock
\APACrefYearMonthDay{2021}{3}{}.
\newblock
{\BBOQ}\APACrefatitle {{Neither measurement error nor speed-accuracy trade-offs
  explain the difficulty of establishing attentional control as a psychometric
  construct: Evidence from a latent-variable analysis using diffusion
  modeling}} {{Neither measurement error nor speed-accuracy trade-offs explain
  the difficulty of establishing attentional control as a psychometric
  construct: Evidence from a latent-variable analysis using diffusion
  modeling}}.{\BBCQ}
\newblock
\APACjournalVolNumPages{PsyArXiv}{}{}{}.
\newblock
\begin{APACrefURL} \url{https://doi.org/10.31234/osf.io/3h26y} \end{APACrefURL}
\PrintBackRefs{\CurrentBib}

\bibitem [\protect \citeauthoryear {%
Ruch%
}{%
Ruch%
}{%
{\protect \APACyear {2020}}%
}]{%
@ruch2020}
\APACinsertmetastar {%
@ruch2020}%
\begin{APACrefauthors}%
Ruch, A.%
\end{APACrefauthors}%
\unskip\
\newblock
\APACrefYearMonthDay{2020}{}{}.
\newblock
{\BBOQ}\APACrefatitle {{Can x2vec save lives? Integrating graph and language
  embeddings for automatic mental health classification}} {{Can x2vec save
  lives? Integrating graph and language embeddings for automatic mental health
  classification}}.{\BBCQ}
\newblock
\APACjournalVolNumPages{Journal of Physics: Complexity}{1}{3}{}.
\PrintBackRefs{\CurrentBib}

\end{thebibliography}

\end{document}